\documentclass[twocolumn]{article}

\usepackage{microtype}
\usepackage{graphicx}
\usepackage{subcaption}
\usepackage{booktabs} 
\usepackage{pgfplots} 
    \pgfplotsset{compat=1.16}
\usepackage{amsmath}
\usepackage{amsthm}
\usepackage{ulem}
\usepackage{svg}
\usepackage[ruled, lined, linesnumbered, commentsnumbered, longend]{algorithm2e}
\usepackage{algorithmic}
\usepackage{natbib}
\usepackage{hyperref}

\theoremstyle{definition}
\newtheorem{defn}{Definition} % definition numbers are dependent on theorem numbers

\title{Adaptive Neighbourhoods for the Discovery of Adversarial Examples}
\author{
Jay Morgan\footnote{Email correspondence can be made to \url{j.p.morgan@swansea.ac.uk}}\\
Department of Computer Science\\
Swansea University, \\
Wales, United Kingdom
\and
Adeline Paiement\\
Université de Toulon, \\
Aix Marseille Univ, \\
CNRS, LIS, Marseille, \\
France
\and
Arno Pauly\\
Department of Computer Science\\
Swansea University, \\
Wales, United Kingdom
\and
Monika Seisenberger\\
Department Computer Science\\
Swansea University, \\
Wales, United Kingdom
}
\date{}

\begin{document}

\maketitle

\begin{abstract}
Deep Neural Networks (DNNs) have often supplied state-of-the-art results in pattern recognition tasks. Despite their advances, however, the existence of adversarial examples have caught the attention of the community. Many existing works have proposed methods for searching for adversarial examples within fixed-sized regions around training points. Our work complements and improves these existing approaches by adapting the size of these regions based on the problem complexity and data sampling density. This makes such approaches more appropriate for other types of data and may further improve adversarial training methods by increasing the region sizes without creating incorrect labels.
\end{abstract}

%%%%%%%%

\section{Introduction}
\label{sec:introduction}

Machine Learning (ML) models are often used in many difficult to define tasks such as image-recognition, language translation, and generation of novel works of art. Deep Neural Networks (DNNs), in particular, are an architectural-type model that has provided many of the state-of-the-art results \citep{Lecun2015}. Despite their impressive performance, DNNs, however, have been shown to be susceptible to \textit{adversarial examples} \citep{Goodfellow-et-al-2016,szegedy2013intriguing}. These examples occur when a small (often human-imperceptible) change to input causes a change in output classification made by the DNN \citep{DBLP:journals/corr/abs-1905-07121}.

\begin{defn}
Given $f$, the classification implemented by a DNN, and some small perturbation $\varepsilon$ of an input $x$, an adversarial $x^{\star}$ is $\varepsilon$-close to $x$ with $f(x^{\star}) \neq f(x)$, while $x^{\star}$ belongs to the same class as $x$.
\end{defn}

It is well studied adversarial examples exist within small regions around the training data. \citet{szegedy2013intriguing} show how the presence of adversarial examples contradicts the general belief that DNN's complexity makes them good at generalising to unseen examples. Later work by \citet{Goodfellow2015} proposes the Fast Gradient Sign Method to generate adversarial examples from a closed $n$-ball around the input. This method perturbs the input pixel values in the direction of the cost function's gradient. While \citet{Goodfellow2015} employ gradient information of the model, \citet{Gu2014TowardsExamples,Huang2017SafetyNetworks} use a black box assumption to find adversarial examples by choosing a random dimension (or pixel) on the input space \cite{Gu2014TowardsExamples} or on activation maps \cite{Huang2017SafetyNetworks} for which an $\varepsilon$ is added and subtracted from the original value. The model is repeatedly queried to determine if the perturbation would result in a misclassification. \citet{Lindenbaum2018GeometryGeneration} apply a diffusion map algorithm to generate a reduced data space. The authors synthesise new points within sparse regions of the data space, under the assumption that more adversarials will occur due to the lack of information in these areas. While we also use the assumption that adversarials may preferably appear in under-sampled areas where DNNs are under-trained, we also consider these areas very carefully, as the location of class boundaries are more uncertain there. Hence, we argue these areas should not be used blindly to search for adversarials, but the areas should be restricted based on uncertainty on class boundary location.

Indeed, while in the context of these studies, the proposed algorithms are targeted at image-based classifiers in which small perturbations don't generally cause a change of class, and can be visually inspected for class type, the same algorithms may pose problems for other types of data. In datasets with jagged class boundaries, small changes may inadvertently push data across true class decision boundaries and thus incorrectly label the data. The focus of our study is therefore to provide a mechanism to quantify the amount of perturbation that can be safely applied (without change of class) to a dataset. This quantification may enable the use of existing adversarial generation algorithms, and allow their use for non-image types of data. Our proposed method estimates the density of samples within the data manifold to identify areas where true class boundaries may be uncertain. This allows defining regions where adversarial generation algorithms may be safely used in future work.

The organisation of this article is as follows: in Section~\ref{sec:related_work} we review the existing material for the characterisation of complexity and density from manifold geometry. Our method is explained in Section~\ref{sec:methodology}. As this is a work in progress, experimental results will be provided in a future publication. We give our concluding remarks in Section~\ref{sec:conclusion}.

\section{Related Work}
\label{sec:related_work}
   
A key element of our study is the determination of the complexity of class boundaries, which impacts the uncertainty on their localisation. \citet{Ho2002ComplexityProblems} examine how 7 different metrics can be used to characterise the complexity in binary classification problems. They are used to form an embedding space where datasets are organised based on their complexity. The metrics include the \textit{Fisher Discriminant Ratio} that considers the distribution of values of a single feature across the elements of a class, and describes the overlap between the distributions of two classes. Later work by \citet{orriols2010documentation} provides an update for the original calculation of the Fisher Discriminant Ratio to account for ordinal features as well as multi-class datasets.

Another metric proposed by \citet{Ho2002ComplexityProblems} is the \textit{Fraction of Hyperspheres Covering Data}. Neighbourhoods are increasingly expanded until they reach a data-point from another class. Smaller neighbourhoods contained within large ones are eliminated and the ratio between the total number of neighbourhoods and the number of data-points is computed. For simpler datasets, a few number of neighbourhoods are needed to cover the dataset. \citet{Lorena2019HowComplexity} provide an additional improvement on this approach by stopping the expansion of neighbourhoods when the spheres from different classes meet. This was used by \citet{Frank1996PretopologicalLearning} as a mean to find the middle line between data points of different classes, in order to estimate the location of class boundaries and hence provide a classification function. Work by \citet{Sinha2019DefendingPertubations} use this notion of covering data to sample possible adversarials to be included in the training procedure for more robustness. Our work builds on these techniques to define adaptive neighbourhoods that incorporate information on the sampling density of the data manifold, making it more applicable for adversarial generation.

More recent work for characterisation of large image datasets is shown by \citet{DBLP:journals/corr/abs-1905-07299}. As image data often resides in a  high-dimensional space, many of the proposed complexity analysis metrics become computationally intractable, especially for large class numbers. A similarity matrix of a lower-dimensional latent representation is created and summarised using spectral clustering. Low eigenvalues for the resulting embedding indicate a low inter-class overlap. They demonstrate this effect on MNIST \citep{LeCun1998Gradient-basedRecognition} where swapped labels increase the eigenvalues. The difference in eigenvalues is normalised and summed to give a final complexity metric.

Many works use properties of the manifold to design methods of improving DNN robustness. \citet{Ahmad2019HowRepresentations} use the sparsity and high-dimensionality to develop sparse DNN weight matrices that appear to increase the overall robustness of the network to random perturbations that may occur due to noise. \citet{Srinivasan2019DefenseDynamics} design an adversarial defence method using Markov Chain Sampling of the manifold. Their technique aims to \textit{drive} adversarial examples towards the more dense regions of the manifold as they explain the output predictions in sparse regions can be unpredictable and more susceptible towards attacks. Our work can be seen as complementary to these approaches by providing a region where the DNNs robustness to adversarial can be appropriately tested.

%%%%%%%%

\section{Methodology}
\label{sec:methodology}

Many approaches for the automated construction of adversarial examples use a fixed-sized $\varepsilon$ for the local neighbourhoods around training points \citep{huang2017adversarial,Goodfellow2015,Goodfellow-et-al-2016}. To ensure these local neighbourhoods are within class boundaries, we propose the use of dataset complexity and of density analysis of the data manifold to provide an adaptive $\varepsilon$ for each sample. 

Our adaptive definition of neighbourhoods relies on the properties of the data manifold, detailed in Section \ref{subsec:manifold_prop}. In particular, they involve the notion of sampling density of the data manifold (Section \ref{subsec:density}) to assist in iteratively building neighbourhoods that may remain within class boundaries according to the available class information (Section \ref{subsec:neighbour}).

\subsection{Manifold properties}
\label{subsec:manifold_prop}

Many ML applications and learning techniques operate under the assumption of a manifold hypothesis \citep{Narayanan2010SampleHypothesis,Brahma2016WhyPerspective}, where real/natural high-dimensional data lie on a low-dimensional manifold embedded within their high-dimensional space. The consequence of this assumption is data has a local homeomorphism with a Euclidean space of lower dimensionality that is a local approximation to the manifold \citep{Brahma2016WhyPerspective}. Therefore, the local neighbourhood of data points may be approximately measured with a Euclidean-based metric.

\begin{defn}
Let $(\mathcal{X},d)$ be a metric space with $\mathcal{X}$ the space of data points $X$ and $d$ the Euclidean distance. The $\varepsilon$-neighbourhood of a point $x_i \in X$ is defined as \citep{Schubert1968Topology}: \[\mathcal{N}_\varepsilon(x_i) := \{x \, | \, d(x,x_i) < \varepsilon \}\]
\end{defn}

Our method considers two properties of the dataset:\\
\textbf{M1} The geometric complexity of the class boundaries.\\
\textbf{M2} The sparsity/density of sampling from the data manifold that constitutes the training data.

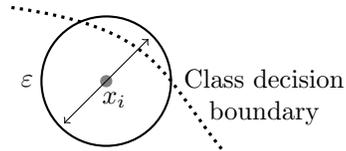
\begin{figure}
    \centering
    \begin{tikzpicture}[scale=0.7]
        \draw [very thick,dotted] (-2,1.2) .. controls (0.8, 0.8) and (0.8,0) .. (2,-1.5);
        \filldraw [gray] (-0.2,-0.2) circle (3pt);
        \draw [->] (-0.5,-0.5) -- (0.6,0.6);
        \draw [->] (-0.5,-0.5) -- (-1,-1);
        \draw [thick] (-0.2,-0.2) circle (35pt);
        
        \node at (-0.05,-0.55) {$x_i$};
        \node at (-1.7, -0.2) {$\varepsilon$};
        \node[align=center] at (2.8, -0.5) {Class decision \\ boundary};
    \end{tikzpicture}
    \caption{Example where a data point $x_i$ lies close to the class decision boundary. In these situations, too large $\varepsilon$ values, may push the synthetically generated point over true class boundaries.}
    \label{fig:complexity}
\end{figure}

\textbf{M1} refers to situations where differently labelled data points lay close together in the topological space, and therefore any perturbation of the data points could result in passing the class boundaries, while wrongly labelling the perturbation the same as the original (Fig.~\ref{fig:complexity}).

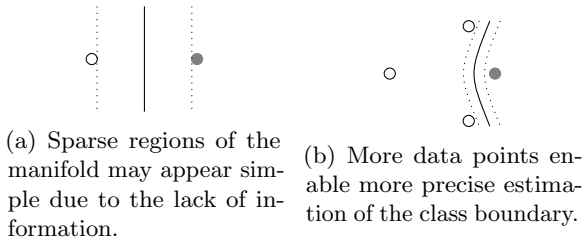
\begin{figure}
    \centering
    \begin{subfigure}{.45\linewidth}
        \centering
        \begin{tikzpicture}[scale=0.7]
            % nodes
            \draw (-1,0) circle (3pt);
            \filldraw [gray] (1,0) circle (3pt);
            
            %lines
            \draw (0, 1) -- (0,-1);
            \draw [dotted] (-0.9,1) -- (-0.9,-1);
            \draw [dotted] ( 0.9,1) -- ( 0.9,-1);
        \end{tikzpicture}
        \caption{Sparse regions of the manifold may appear simple due to the lack of information.}
        \label{fig:density_a}
    \end{subfigure}
    \;
    \begin{subfigure}{.45\linewidth}
        \centering
        \begin{tikzpicture}[scale=0.7]
            % nodes
            \draw (-1,0) circle (3pt);
            \draw (0.5,0.9) circle (3pt);
            \draw (0.5,-0.9) circle (3pt);
            \filldraw [gray] (1,0) circle (3pt);
            
            %lines
            \draw (0.9, 1) .. controls (0.5,0) .. (0.9,-1);
            \draw [dotted] (1.1, 1) .. controls (0.7,0) .. (1.1,-1);
            \draw [dotted] (0.7, 1) .. controls (0.3,0) .. (0.7,-1);
        \end{tikzpicture}
        \caption{More data points enable more precise estimation of the class boundary.}
        \label{fig:density_b}
    \end{subfigure}
    \caption{Example scenario where true class boundaries are revealed when more data is collected.}
    \label{fig:density}
\end{figure}

\textbf{M2} concerns the number of samples from different regions of the data manifold. In sparse regions (small numbers of samples), estimated class boundaries may seem deceivingly simple, e.g. linear with a wide margin \citep{Ho2002ComplexityProblems} (Fig.~\ref{fig:density_a}). By increasing the number of examples (i.e. collecting more data), the true complexity of the classification task may become apparent (Fig.~\ref{fig:density_b}).

\subsection{Estimating sparsity/density}
\label{subsec:density}

Sparsity of the manifold is measured using KDE from support examples for each class. KDE uses a real-valued radial basis function (RBF) $\varphi$, a function of the distance between some point $\overline{x}$ (the centre or origin) and another point $x$. We use the inverse-multiquadric function (Eq.~\ref{eq:rbf}) as it has a non-shrinking value away from the origin. 
\begin{equation}
    \varphi(x; \overline{x}) =  \frac{1}{\sqrt{1 + (\varepsilon r)^2}},\; \text{where}\; r = \parallel \overline{x} - x \parallel
    \label{eq:rbf}
\end{equation}

Providing the RBF's width parameter is suitably chosen, we achieve a good measure of the density through the sum of the RBFs centred on all data points $X^c$ of class $c$ (Eq.~\ref{eq:density}).
\begin{equation}
    \rho_c(x) = \sum_{x_j \in X^c} \varphi(x; x_j)
    \label{eq:density}
\end{equation}
% \sout{where $\mathcal{X}^c$ is the set of inputs of class $c$ \apn{$x$ is not really a point with a class, it is a possibility: any possible point $x$ in the KDE's input space} \jmn{is this a comment?}.}

\subsection{Constructing neighbourhoods}
\label{subsec:neighbour}

We construct an adaptive neighbourhood for each point of our dataset, as a sphere of finely tuned radius. The neighbourhoods for all data points are created by jointly maximising the individual volumes of the spheres, under the constraint that sphere of different classes don't overlap. In addition, to account for lack of knowledge in under-sampled areas, we keep the sphere's volume limited to a linear function of the local sampling density $\rho_c(x)$ for the class.
This is expressed by the following Lagrangian function:
\begin{equation}
\begin{split}
    L(\varepsilon_1,\cdots,\varepsilon_n,x_1,\cdots,x_n,\lambda)=v(\varepsilon_1,\cdots,\varepsilon_n) \\ +\lambda_g g(\varepsilon_1,\cdots,\varepsilon_n,x_1,\cdots,x_n) \\ +\lambda_h h(\varepsilon_1,\cdots,\varepsilon_n,x_1,\cdots,x_n)
\end{split}
\end{equation}
where $v$ is the volume function, from $R^n$ to $R^n$, to be maximised:
\begin{equation}
    v(\varepsilon_1,\cdots,\varepsilon_n) = \begin{pmatrix}
        \varepsilon_1^D \\ \vdots \\ \varepsilon_{n}^D
    \end{pmatrix}
\end{equation}
and
\begin{multline}
    g(\varepsilon_1,\cdots,\varepsilon_n,x_1,\cdots,x_n) = \\ \begin{pmatrix} \sum_{\substack{j \neq 1\\
    c(j)\neq c(1)}} \min(d\left(x_1,x_j\right) - (\varepsilon_1 + \varepsilon_j), 0) \\
    \vdots \\
    \sum_{\substack{j\neq n\\
    c(j)\neq c(n)}} \min(d\left(x_n,x_j\right) - (\varepsilon_n + \varepsilon_j), 0)
    \end{pmatrix}
        \label{eq:intersection}
\end{multline}
\begin{multline}
    h(\varepsilon_1,\cdots,\varepsilon_n,x_1,\cdots,x_n)= \\ \begin{pmatrix}
        \varepsilon_1^D \\ \vdots \\ \varepsilon_{n}^D
    \end{pmatrix} - \alpha \begin{pmatrix}
        \rho_{c(1)}(x_1) \\ \vdots \\ \rho_{c(n)}(x_n)
    \end{pmatrix} + \beta
    \label{eq:local_density_constraint}
\end{multline}
are the Lagrangian constraints for no intersection and volume depending linearly on density, respectively, which should be both equal to zero. $c(i)$ is the class of point $i$.
Note that when two spheres of different classes are too close to each other, they may not simultaneously respect both constraints of not intersecting while attaining their full size depends on local density for their respective classes. Therefore, the optimisation problem needs to be relaxed. 

\begin{figure}
    \centering
    \begin{tikzpicture}[scale=0.6]
        \draw (0.4,0) node {$x_1$};
        \draw[dashed] (0,0) circle (1.0cm);
        \draw[dashed] (0,0) circle (1.45cm);
        \draw[thick,dotted] (0,0) circle (1.75cm);
        
        \draw[->]        (0.1,0) -- (-1.0,0) node[below,midway] {$\varepsilon_1$};
        \draw[->] (-1.0,0) -- (-1.45,0) node[below,midway] {};
        \draw[->] (-1.5,0) -- (-1.75,0) node[below,midway] {};
        \draw[thick, ->] (0.1,0) -- (-0.5,1.75) node[anchor=south] {$\varepsilon$};
        
        \draw (2.57,1) node {$x_2$};
        \draw[thick] (2.57,1) circle (1.0cm);
        
        \draw (2,-0.4) node {$x_3$};
        \draw[thick,dotted] (2,-0.4) circle (0.5cm);
    \end{tikzpicture}
\caption{Iterative $\varepsilon$-expansion process in a binary class scenario. The two classes are distinguished by the dotted and solid circles.}
\label{fig:e_expansion}
\end{figure}
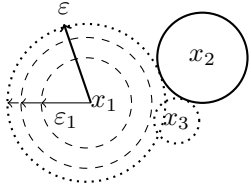

An iterative algorithm may achieve an approximate result as the relaxed optimisation (Fig.~\ref{fig:e_expansion}). This iterative version is reminiscent of classification algorithms by \citet{Frank1996PretopologicalLearning}, and complexity analysis algorithms by \citet{Ho2002ComplexityProblems} and \citet{Lorena2019HowComplexity}, but further developed to incorporate a decay function to limit the expansion of the neighbourhoods based on local density.

% \begin{equation}
%     \mathcal{I}_{x_i} = \sum_{c(x_i) \neq c(x_j)} \min(|| x_i - x_j || - \epsilon_i + \epsilon_j, 0)
%     \label{eq:intersection}
% \end{equation}
% \apn{This equation seems to be the same as a term of $g$. Maybe we don't need to repeat its definition?}

% \sout{\ap{Our iterative algorithm accounts for the constraint of} \sout{In order to enforce} no intersection of neighbourhoods of differing classes \ap{expressed in}\sout{, the intersection between a data point of one class and all points of different classes are summed (}Eq. \ref{eq:intersection}\sout{). The result of this sum shall be 0 when there are no intersections} \ap{which is optimised to a 0 value}.}

Small initial neighbourhoods are progressively expanded, with their radius at iteration $n$ being $\varepsilon_i^n=\varepsilon_i^{n-1}+\Delta\varepsilon_i^n$, subject to avoiding overlap of neighbourhoods from different classes (Eq.~\ref{eq:intersection}), and with an exponentially decreasing expansion that further depends on the local density of samples for the related class (Eq.~\ref{eq:local_density_constraint}):
\begin{equation}
    \Delta\varepsilon_i^n=e^{-\rho_{c(i)}(x_i) \cdot n}
    \label{eq:step}
\end{equation}
In areas of low density, so with insufficient number of samples to safely determine the location of class boundaries, the expansion is slower and generates a conservative small final neighbourhood. Expansion stops when it reaches a low threshold $\Delta\varepsilon^{min}$ making it insignificant.

\begin{algorithm}[h]
    \caption{Calculate $\varepsilon_i$ for data point $x_i$}
    \label{alg:epsilon_expansion}
    
    \SetKwInOut{KwIn}{Input}
    \SetKwInOut{KwOut}{Output}
    
    \KwIn{each sample of manifold $X$}
    \KwOut{$\varepsilon$ value for each sample of $X$}

    $\Delta\varepsilon^{min} \gets 1e-20$ \\
    
    \For{$x_i \in X$}{
        $\varepsilon_i \gets 0$\\
        $stop_i \gets false$
    }
    
    %  $n \gets 1$
    
    \While{$\exists i \text{ such that } stop_i = false$}{
            \For{$x_i \in X$ such that $stop_i = false$}{
                \For{$x_j \not\in X^c(i)$}{
                    \If{$ d(x_i, x_j) \leq \varepsilon_i + \varepsilon_j$}{
                       $stop_i \gets true$
                    }
                }
                \If{$stop_i = false$}{
                    $\Delta\varepsilon_i \gets e^{-\rho_{c(i)}(x_i)n}$\\
                    \If{$\Delta\varepsilon_i \leq \Delta\varepsilon^{min}$}{
                        $stop_i \gets true$
                    }
                    \Else{
                        $\varepsilon_i \gets \varepsilon_i + \Delta\varepsilon_i$
                    }
                }
            }
            % $n \gets n + 1$
        }
    \KwRet{$\varepsilon$}
\end{algorithm}

This method may be further improved in future work by accounting for the complexity of class boundaries in Eq.~\ref{eq:step}, e.g. using complexity metrics of \citet{Ho2002ComplexityProblems}.

\begin{figure}
    \centering
    \includegraphics[width=\linewidth]{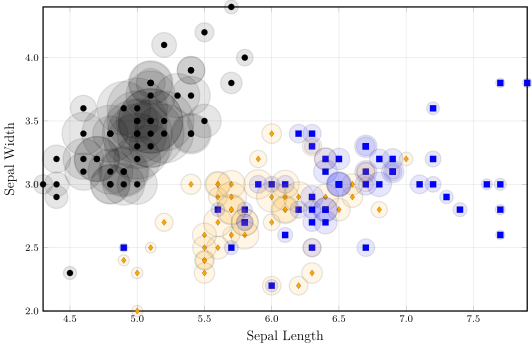}
    \caption{Proposed adaptive neighbourhoods for the Iris dataset. The three classes of flower are represented by different shaped markers. The size of the neighbourhood for each sample is indicated with a circle centred on the data point. Intersections between neighbourhoods of different classes are not real but are visualisation artefacts coming from the 2D projection of 4 dimensions.}
    \label{fig:iris}
\end{figure}

We provide results for the Iris flower dataset \citep{Fisher1936TheProblems} in Fig~\ref{fig:iris}. In regions containing densely-packed samples of the same class, neighbourhoods tend to grow larger and cover the space as there is more information to be more confident about class boundary placement. In other sparse regions with uncertain location of class boundaries however, neighbourhood size is limited appropriately. The use of this dataset aims to demonstrate the possible support from our algorithm for adversarial generation for non image-based datasets. Future work will consider the effectiveness of existing adversarial training algorithms when amounts of perturbation are specified by these neighbourhoods.

%%%%%%%%%

\section{Conclusion}
\label{sec:conclusion}

We propose a method to characterise the amount of perturbation that can be safely applied to data without a change of class label, in order to ease the search for adversarials. It uses two properties of the data manifold to address two main concerns with current automatic generation of adversarial examples: (1) sparse regions of the manifold does not give enough information as to the true class boundaries, and blind perturbations may push data points across these boundaries; and (2) a single value of perturbation for the entire dataset does not appropriately cover the geometric complexities of the class boundaries.

In this article, we demonstrate an iterative method to determine adaptive sizes of neighbourhoods based on local sampling density. These neighbourhoods may provide a search space for existing algorithms to generate adversarial examples. Moreover, our method may further enable the use of existing adversarial training algorithms for non image-based datasets. Our method may be further improved by accounting for the complexity of the classification problem, and therefore of class boundaries, in the design of our adaptive neighbourhoods. This improvement and further experiments are left for future work.

% \sout{\ap{In future works, our method may be further improved by accounting for the complexity of classification problem, and therefore of class boundaries, in the design of our adaptive neighbourhoods.}}

%%%%%%%%%%%

% \section{Conclusion}
% \label{sec:conclusion}

% We have demonstrated a method to quantify the safe amount of perturbation that can be made to input data, without passing estimated class boundaries. Our proposed algorithm works by jointly maximising the size of open neighbourhoods centred on the data points, while minimising the class overlap. This optimisation takes into consideration the manifold properties to which it is operating upon. 

% This \jm{ongoing} work \sout{can} \jm{may} be used to improve existing algorithms for searching and generating adversarial examples during training. \jm{This is left for future work}.

\bibliography{main}
\bibliographystyle{plainnat}
\end{document}